\documentclass[11pt,a4paper]{article}
\usepackage[hyperref]{acl2021}
\usepackage{times}
\usepackage{latexsym}
\UseRawInputEncoding

\usepackage{microtype}

\usepackage{subfiles}
\usepackage{amsmath}
\usepackage{multirow}
\usepackage{graphicx}
\usepackage{xcolor}
\usepackage[boxed]{algorithm2e}
\usepackage{here}
\usepackage{amssymb}

\usepackage{lineno}

\newcommand{\argmin}{\mathop{\rm argmin}\limits}
\newcommand{\vect}[1]{\mathbf{{#1}}}

\aclfinalcopy 
\def\aclpaperid{88} 

\title{Data Augmentation with Unsupervised Machine Translation Improves \\ the Structural Similarity of Cross-lingual Word Embeddings}

\author{Sosuke Nishikawa, Ryokan Ri and Yoshimasa Tsuruoka \\
  The University of Tokyo \\
  7-3-1 Hongo, Bunkyo-ku, Tokyo, Japan \\
  {\tt sosuke-nishikawa@nii.ac.jp} \\
  {\tt \{li0123,tsuruoka\}@logos.t.u-tokyo.ac.jp} \\
  }

\date{}

\begin{document}
\maketitle
\begin{abstract}
Unsupervised cross-lingual word embedding (CLWE) methods learn a linear transformation matrix that maps two monolingual embedding spaces that are separately trained with monolingual corpora.
This method relies on the assumption that the two embedding spaces are structurally similar, which does not necessarily hold true in general.
In this paper, we argue that using a pseudo-parallel corpus generated by an unsupervised machine translation model facilitates the structural similarity of the two embedding spaces and improves the quality of CLWEs in the unsupervised mapping method.
We show that our approach outperforms other alternative approaches given the same amount of data, and, through detailed analysis, we show that data augmentation with the pseudo data from unsupervised machine translation is especially effective for mapping-based CLWEs because (1) the pseudo data makes the source and target corpora (partially) parallel; (2) the pseudo data contains information on the original language that helps to learn similar embedding spaces between the source and target languages.
\end{abstract}


\section{Introduction}
\label{intro}

Cross-lingual word embedding (CLWE) methods aim to learn a shared meaning space between two languages (the source and target languages), which is potentially useful for cross-lingual transfer learning or machine translation \cite{yuan-etal-2020-interactive-refinement, Artetxe:2018ta,Lample:2018ts}.
Although early methods for learning CLWEs often utilize multilingual resources such as parallel corpora \cite{Gouws:2015ws,Luong:2015uh} and word dictionaries \cite{Mikolov:2013tp}, recent studies have focused on fully unsupervised methods that do not require any cross-lingual supervision \cite{lample2018word,Artetxe:2018vx,Patra:2019tq}.
Most unsupervised methods fall into the category of mapping-based methods, which generally consist of the following procedures: train monolingual word embeddings independently in two languages; then, find a linear mapping that aligns the two embedding spaces.
The mapping-based method is based on a strong assumption that the two independently trained embedding spaces have similar structures that can be aligned by a linear transformation, which is unlikely to hold true when the two corpora are from different domains or the two languages are typologically very different \cite{Sogaard:2018vo}.
To address this problem, several studies have focused on improving the structural similarity of monolingual spaces before learning mapping \cite{zhang-etal-2019-girls, vulic-etal-2020-improving}, but few studies have focused on how to leverage the text data itself.

\begin{figure}[h]
  \centering
  \includegraphics[width=5.5cm]{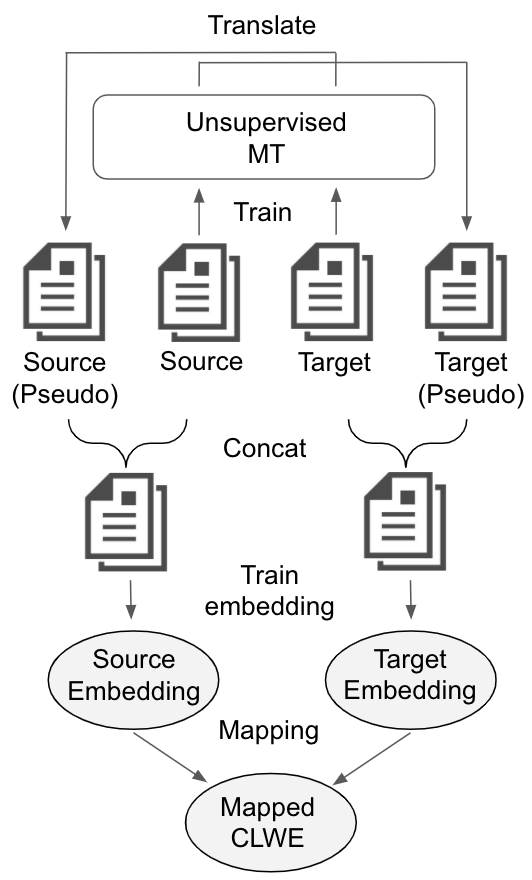}
  \caption{Our framework for training CLWEs using unsupervised machine translation (UMT). We first train UMT models using monolingual corpora for each language. We then translate all the training corpora and concatenate the outputs with the original corpora, and train monolingual word embeddings independently. Finally, we map these word embeddings on a shared embedding.}
  \label{our_framework}
\end{figure}

In this paper, we show that the pseudo sentences generated from an unsupervised machine translation (UMT) system \citep{lample-etal-2018-phrase} facilitates the structural similarity without any additional cross-lingual resources.
In the proposed method, the training data of the source and/or target language are augmented with the pseudo sentences (Figure \ref{our_framework}).

We argue that this method facilitates the structural similarity between the source and target embeddings for the following two reasons.
Firstly, the source and target embeddings are usually trained on monolingual corpora. The difference in the content of the two corpora may accentuate the structural difference between the two resulting embedding spaces, and thus we can mitigate that effect by making the source and target corpora parallel by automatically generated pseudo data.
Secondly, in the mapping-based method, the source and target embeddings are trained independently without taking into account the other language. Thus, the embedding structures may not be optimal for CLWEs. We argue that pseudo sentences generated by a UMT system contain some trace of the original language, and using them when training monolingual embeddings can facilitate the structural correspondence of the two sets of embeddings.

In the experiments using the Wikipedia dump in English, French, German, and Japanese, we observe substantial improvements by our method in the task of bilingual lexicon induction and downstream tasks without hurting the quality as monolingual embeddings.
Moreover, we carefully analyze why our method improves the performance, and the result confirms that making the source and target corpora parallel does contribute to performance improvement, and also suggests that the generated translation data contain information about the original language.


\section{Background and Related Work}
\label{sec:BR}

\subsection*{Cross-lingual Word Embeddings}
CLWE methods aim to learn a semantic space shared between two languages.
Most of the current approaches fall into two types of methods: joint-training approaches and mapping-based approaches.

Joint-training approaches jointly train a shared embedding space given multilingual corpora with cross-lingual supervision such as parallel corpora \cite{Gouws:2015ws,Luong:2015uh}, document-aligned corpora \cite{Vulic:2016tj}, or monolingual corpora along with a word dictionary \cite{Duong:2016vw}.

On the other hand, mapping-based approaches utilize monolingual embeddings that are already obtained from monolingual corpora.
They assume structural similarity between monolingual embeddings of different languages and attempt to obtain a shared embedding space by finding a transformation matrix $\vect{W}$ that maps source word embeddings to the target embedding space \cite{Mikolov:2013tp}.
The transformation matrix $\vect{W}$ is usually obtained by minimizing the sum of squared euclidian distances between the mapped source embeddings and target embeddings:

\begin{equation}
    \label{mse}
    \argmin_{\vect{W}} \sum_{i}^{|D|} \left\|\vect{W} \vect{x}_i - \vect{y}_i \right\|^{2},
\end{equation}

\noindent where $D$ is a bilingual word dictionary that contains word pairs $(x_i, y_i)$ and $\vect{x}_i$ and $\vect{y}_i$ represent the corresponding word embeddings.


Although finding the transformation matrix $\vect{W}$ is straightforward when a word dictionary is available, a recent trend is to reduce the amount of cross-lingual supervision or to find $\vect{W}$ in a completely unsupervised manner \cite{lample2018word,Artetxe:2018vx}.
The general framework of unsupervised mapping methods is based on heuristic initialization of a seed dictionary $D$ and iterative refinement of the transformation matrix $\vect{W}$ and the dictionary $D$, as described in Algorithm \ref{unsup_algo}.
In our experiment, we use the unsupervised mapping-based method proposed by \citet{Artetxe:2018vx}. Their method is characterized by the seed dictionary initialized with nearest neighbors based on similarity distributions of words in each language.


\begin{algorithm*}


\SetKwInput{KwInput}{Input}
\SetKwInput{KwOutput}{Output}
\SetAlgoLined

\caption{The general workflow of unsupervised mapping methods}
\label{unsup_algo}

\KwInput{The source embeddings $\vect{X}$, the target embeddings $\vect{Y}$}
\KwOutput{The transformation matrix $\vect{W}$}
Heuristically induce an initial seed word dictionary $D$

\While{not convergence}{
  Compute $\vect{W}$ given the word dictionary $D$ from the equation (\ref{mse})

  Update the word dictionary $D$ by retrieving cross-lingual nearest neighbors in a shared embedding space obtained by $\vect{W}$}
  \Return $\vect{W}$

\end{algorithm*}

These mapping-based methods, however, are based on the strong assumption that the two independently trained embedding spaces have similar structures that can be aligned by a linear transformation.
Although several studies have tackled improving the structural similarity of monolingual spaces before learning mapping \cite{zhang-etal-2019-girls, vulic-etal-2020-improving}, not much attention has been paid to how to leverage the text data itself.



In this paper, we argue that we can facilitate structural correspondence of two embedding spaces by augmenting the source or/and target corpora with the output from an unsupervised machine translation system \citep{lample-etal-2018-phrase}.

\subsection*{Unsupervised Machine Translation}
Unsupervised machine translation (UMT) is the task of building a translation system without any parallel corpora \cite{Artetxe:2018ta,Lample:2018ts,lample-etal-2018-phrase,artetxe-etal-2019-effective}. UMT is accomplished by three components: (1) a word-by-word translation model learned using unsupervised CLWEs; (2) a language model trained on the source and target monolingual corpora; (3) a back-translation model where the model uses input and its own translated output as parallel sentences and learn how to translate them in both directions.

More specifically, the initial source-to-target translation model $P^{0}_{s \rightarrow t}$ is created by the word-by-word translation model and the language model of the target language.
Then, $P^{1}_{t \rightarrow s}$ is learned in a supervised setting using the source original monolingual corpus paired with the synthetic parallel sentences of the target language generated by $P^{0}_{s \rightarrow t}$. Again, another source-to-target translation model $P^{1}_{s \rightarrow t}$ is trained with the target original monolingual corpus and the outputs of $P^{0}_{s \rightarrow t}$, and in the same way, the quality of the translation models is improved with an iterative process.


In our experiments, we adopt an unsupervised phrase-based statistical machine translation (SMT) method to generate a pseudo corpus
because it produces better translations than unsupervised neural machine translation on low-resource languages \cite{lample-etal-2018-phrase}. The difference of the unsupervised SMT (USMT) model from its supervised counterpart is that the initial phrase table is derived based on the cosine similarity of unsupervised CLWEs, and the translation model is iteratively improved by pseudo parallel corpora.

Our proposed method utilizes the output of a USMT system to augment the training corpus for CLWEs.

\subsection*{Exploiting UMT for Cross-lingual Applications}

There is some previous work on how to use UMT to induce bilingual word dictionaries or improve CLWEs.
\citet{artetxe-etal-2019-bilingual} explored an effective way of utilizing a phrase table from a UMT system to induce bilingual dictionaries.
 \citet{marie-fujita-2019-unsupervised} generate a synthetic parallel corpus from a UMT system, and jointly train CLWEs along with the word alignment information \cite{Luong:2015uh}.
In our work, we use the synthetic parallel corpus generated from a UMT system not for joint-training but for data augmentation to train monolingual word embeddings for each language, which are subsequently aligned through unsupervised mapping. 
In the following sections, we empirically show that our approach leads to the creation of improved CLWEs and analyze why these results are achieved.


\section{Experimental Design}

\label{sec:ED}

In this section, we describe how we obtain mapping-based CLWEs using a pseudo parallel corpus generated from UMT.
We first train UMT models using the source/target training corpora, and then translate them to the machine-translated corpora. Having done that, we simply concatenate the machine-translated corpus with the original training corpus, and learn monolingual word embeddings independently for each language. Finally, we map these embeddings to a shared CLWE space.

\subsection*{Corpora}

We implement our method with two similar language pairs: English-French (en-fr), English-German (en-de), and one distant language pair: English-Japanese (en-ja). 
We use plain texts from Wikipedia dumps\footnote{\url{https://dumps.wikimedia.org/}}, and randomly extract 10M sentences for each language.
The English, French, and German texts are tokenized with the Moses tokenizer \cite{koehn-etal-2007-moses} and lowercased. For Japanese texts, we use {\tt kytea}\footnote{\url{http://www.phontron.com/kytea/index-ja.html}} to tokenize and normalize them\footnote{We convert all alphabets and numbers to half-width, and all katakana to full-width with the {\tt mojimoji} library \url{https://github.com/studio-ousia/mojimoji}}.

\subsection*{Training mapping-based CLWEs}

Given tokenized texts, we train monolingual word embeddings using {\tt fastText}\footnote{\url{https://fasttext.cc}} with 512 dimensions, a context window of size 5, and 5 negative examles.
We then map these word embeddings on a shared embedding space using the open-source implementation {\tt VecMap}\footnote{\url{https://github.com/artetxem/vecmap}} with the unsupervised mapping algorithm \cite{Artetxe:2018vx}.

\subsection*{Training UMT models}

To implement UMT, we first build a phrase table by selecting the most frequent 300,000 source phrases and taking their 200 nearest-neighbors in the CLWE space following the setting of \citet{lample-etal-2018-phrase}.
We then train a 5-gram language model for each language with {\tt KenLM} \cite{heafield-etal-2013-scalable} and combine it with the phrase table, which results in an unsupervised phrase-based SMT model.
Then, we refine the UMT model through three iterative back-translation steps. At each step, we translate 100k sentences randomly sampled from the monolingual data set. We use a phrase table containing phrases up to a length of 4 except for initialization.
The quality of our UMT models is indicated by the BLEU scores \cite{DBLP:conf/acl/PapineniRWZ02} in Table \ref{table:blue_results}. We use newstest2014 from WMT14\footnote{\url{http://www.statmt.org/wmt14/translation-task.html}} to evaluate En-Fr and En-De translation accuracy and the Tanaka corpus\footnote{\url{http://www.edrdg.org/wiki/index.php/Tanaka Corpus}} for En-Ja evaluation.

\begin{table}[t]
    \begin{center}
     \scalebox{0.7}{
      \begin{tabular}{cccccc} \hline
       \multicolumn{2}{c}{en - fr} & \multicolumn{2}{c}{en - de} & \multicolumn{2}{c}{en - ja}\\ \hline 
       $ \rightarrow $ & $ \leftarrow $ & $ \rightarrow $ & $ \leftarrow $ & $ \rightarrow $& $ \leftarrow $\\ \hline \hline
         19.2 & 19.1& 10.3& 13.7 & 3.6 & 1.4
        \\ \hline
      \end{tabular}
      }
      \caption{BLEU scores of UMT.}
      \label{table:blue_results}
    \end{center}

  \end{table}
  
\subsection*{Training CLWEs with pseudo corpora}

We then translate all the training corpora with the UMT system and obtain machine-translated corpora, which we call {\it pseudo corpora}.
We concatenate the pseudo corpora with the original corpora, and learn monolingual word embeddings for each language.
Finally, we map these word embeddings to a shared CLWE space with the unsupervised mapping algorithm.

\subsection*{Models}
We compare our method with a baseline with no data augmentation as well as the existing related methods: dictionary induction from a phrase table \cite{artetxe-etal-2019-bilingual} and the unsupervised joint-training method \cite{marie-fujita-2019-unsupervised}. These two methods both exploit word alignments in the pseudo parallel corpus, and to obtain them we use {\tt Fast\_Align}\footnote{\url{https://github.com/clab/fast_align}} \cite{dyer-etal-2013-simple} with the default hyperparameters. 
For the joint-training method, we adopt {\tt bivec}\footnote{\url{https://github.com/lmthang/bivec}} to train CLWEs with the parameters used in \citet{upadhyay-etal-2016-cross} using the pseudo parallel corpus and the word alignments.
To ensure fair comparison, we implement all of these methods with the same UMT system.

\begin{table*}[t]
  \begin{center}
    \scalebox{0.65}{
    \begin{tabular}{c|cccc|cc|cc|cc|cc|cc|ccc}
     \hline
        \multirow{2}{*}{Method} & \multicolumn{2}{c}{source (en)} & \multicolumn{2}{c}{target}  &  \multicolumn{2}{|c|}{en$\rightarrow$fr} &  \multicolumn{2}{|c|}{fr$\rightarrow$en}&  \multicolumn{2}{|c|}{en$\rightarrow$de}&  \multicolumn{2}{c|}{de$\rightarrow$en} &  \multicolumn{2}{|c|}{en$\rightarrow$ja}& \multicolumn{2}{c}{ja$\rightarrow$en}\\ 
        
　　　　　 & orig. & psd. & orig. & psd.  & MRR& P@1 & MRR& P@1&  MRR& P@1&  MRR& P@1 &  MRR& P@1& MRR& P@1\\
        
        \hline \hline

      \multirow{3}{*}{\begin{tabular}[c]{@{}c@{}c@{}c@{}}BLI from\\phrase table\end{tabular}}  & \checkmark & - & -  & \checkmark & -&$\textrm{0.673} $& -  &$\textrm{0.524} $& - &
      $\textrm{0.551}$&  - & $\textrm{0.486}$&   - & $\textrm{0.311}$ & - & $\textrm{0.226} $
      \\ \cline{2-17}
     & - & \checkmark & \checkmark & -& - &$\textrm{0.509} $ &-  & $\textrm{0.697} $& - &$\textrm{0.302}$ &-& $\textrm{0.542}  $& - & $\textrm{0.198}$ &
       - & $\textrm{0.259} $
      \\\cline{2-17}
     & \checkmark & \checkmark & \checkmark & \checkmark &-&$\textrm{0.673} $& - & $\textrm{0.522} $& -& $\textrm{0.551}$ &- & $\textrm{0.486}  $&-&
     $ \textrm{0.311}  $ &  - & $\textrm{0.226} $

      \\ \hline

      \multirow{3}{*}{ \begin{tabular}[c]{@{}c@{}}joint\\training\end{tabular}} & \checkmark & - & - & \checkmark& $\textrm{0.640} $ &$\textrm{0.636} $& $\textrm{0.615} $ &$\textrm{0.634} $ & $\textrm{0.552} $  & $\textrm{0.509}$ & $\textrm{0.545} $ &$\textrm{0.520}$& $\textrm{0.347}$ &0.295  & $\textrm{0.272}$ & $\textrm{0.227} $
      \\ \cline{2-17}
      & - & \checkmark & \checkmark & - & $\textrm{0.587}  $&$\textrm{0.579} $& $\textrm{0.643} $  &$\textrm{0.685} $& $\textrm{0.535} $ &$\textrm{0.491}  $&  $\textrm{0.577} $ &$\textrm{0.549}  $& $\textrm{0.279} $&0.226  & $\textrm{0.305}$ & $\textrm{0.249} $
      \\ \cline{2-17}
    & \checkmark & \checkmark & \checkmark & \checkmark & $\textrm{0.654} $& $\textrm{0.642} $& $\textrm{0.642} $  &$\textrm{0.650} $& $\textrm{0.585}$ &$\textrm{0.532}  $&  $\textrm{0.520}  $ &$\textrm{0.518}  $& $\textrm{0.325}   $ & 0.267& $\textrm{0.295}$&
      $\textrm{0.234} $
      \\\hline

         mapping& \checkmark & - & \checkmark & -  & $\textrm{0.670}   $  &  $\textrm{0.612}   $ & $\textrm{0.650} $  &$\textrm{0.614} $& $\textrm{0.579} $ & $\textrm{0.484}$ & $\textrm{0.587} $  &$\textrm{0.488}$& $\textrm{0.471}$ & 0.378   & $\textrm{0.364} $ & 0.242
       \\ \hline

       \multirow{3}{*}{\begin{tabular}[c]{@{}c@{}}mapping\\(+ pseudo)\end{tabular}}   & \checkmark &- & \checkmark & \checkmark  & $\textrm{0.709}  $  & $\textrm{0.666}   $ & $\textrm{0.687} $  &$\textrm{0.688} $& $\textbf{0.656}$
        & $\textbf{0.582}$ & $\textrm{0.635}$   &$\textbf{0.563}$&$\textbf{0.514} $& $\textbf{0.405 }$  & $\textbf{0.436} $ & $\textbf{0.304} $
      \\ \cline{2-17}
     & \checkmark & \checkmark & \checkmark & -  & $\textbf{0.728} $  & $\textbf{0.684}   $ & $\textbf{0.703} $ &$\textbf{0.700} $ & $\textrm{0.647}$ & $\textrm{0.566}$ &$\textrm{0.636}$  &$\textrm{0.562}$& $\textrm{0.486}$ &0.392   & $\textrm{0.407} $ & $\textrm{0.297} $
      \\\cline{2-17}
  & \checkmark & \checkmark & \checkmark & \checkmark  & $\textrm{0.721} $  & $\textrm{0.677}   $ & $\textrm{0.696} $ &$\textbf{0.700} $ & $\textrm{0.652} $ & $\textrm{0.574}$ &$\textbf{0.637} $ &$\textbf{0.563}$ & $\textrm{0.497}$&  0.387  & $\textrm{0.426} $ & $\textrm{0.300} $
         \\ \hline

    \end{tabular}
    }

    \caption{Comparison with previous approaches in BLI. ``orig." and ``psd." indicate original training corpus and pseudo corpus. In each cell, the left cell shows the result of MRR, and the right cell shows the result of p@1.}
    \label{table:bli_result}

  \end{center}
\end{table*}


\section{Evaluation of Cross-lingual Mapping}
\label{sec:ECM}

In this section, we conduct a series of experiments to evaluate our method. We first evaluate the performance of cross-lingual mapping in our method  (§ \ref{subsec: bli}) and investigate the effect of UMT quality (§ \ref{subsec: umt_quality}). Then, we analyze why our method improves the bilingual lexicon induction (BLI) performance. Through carefully controlled experiments, we argue that it is not simply because of data augmentation but because: (1) the generated data makes the source and target corpora (partially) parallel (§ \ref{subsec: content}); (2) the generated data reflects the co-occurrence statistics of the original language (§ \ref{subsec: co-occurrence}).

\subsection{Bilingual Lexicon Induction}
\label{subsec: bli}

First, we evaluate the mapping accuracy of word embeddings using BLI.
BLI is the task of identifying word translation pairs, and is a common benchmark for evaluating CLWE methods.
In these experiments, we use Cross-Domain Similarity Local Scaling \cite{lample2018word} as the method for identifying translation pairs in the two embedding spaces. For BLI scores, we adopt the mean reciprocal rank (MRR) \cite{glavas-etal-2019-properly} and P@1.

We use XLing-Eval\footnote{\url{https://github.com/codogogo/xling-eval}} as test sets for En-Fr and En-Ge. For En-Ja. We create the word dictionaries automatically using Google Translate\footnote{\url{https://translate.google.com/}}, following \citet{ri-tsuruoka-2020-revisiting}.
Other than BLI from a phrase table, we train three sets of embeddings with different random seeds and report the average of the results.

We compare the proposed method with other alternative approaches in BLI as shown in Table \ref{table:bli_result}. 
In all the language pairs, the mapping method with pseudo data augmentation achieves better performance than the other methods.
Here, one may think that the greater amount of data can lead to better performance, and thus augmenting both the source and target corpora shows the best performance. However, the result shows that it is not necessarily the case: for our mapping method, augmenting only either the source or target, not both, achieves the best performance in many language pairs.
This is probably due to the presence of two pseudo corpora with different natures.

As for the two methods using word alignments (BLI from phrase table; joint training), we observe some cases where these models underperform the mapping methods, especially in English and Japanese pairs.
We attribute this to our relatively low-resource setting where the quality of the synthetic parallel data is not sufficient to perform these methods which require word alignment between parallel sentences.


\begin{table}[t]
    \begin{center}
     \scalebox{0.65}{
      \begin{tabular}{p{1.5em}|p{1.5em}p{1.5em}p{2.5em}|p{1.5em}p{1.5em}p{2.5em}} \hline
        &\multicolumn{3}{c}{en - fr } & \multicolumn{3}{c}{ en - de }\\ \hline 
      BT   &\multicolumn{2}{c}{BLI} &  \multirow{2}{*}{BLEU} & \multicolumn{2}{c}{BLI} &  \multirow{2}{*}{BLEU} \\
       step  &MRR & P@1 &  & MRR & P@1 & \\ \hline \hline 
        \multicolumn{1}{c|}{-} & 0.670 & 0.612 & \multicolumn{1}{c|}{-}  & 0.579 & 0.484 &  \multicolumn{1}{c}{-}   \\ \hline 
        \multicolumn{1}{c|}{0} &  0.711 & 0.646 & \multicolumn{1}{c|}{14.7}  & 0.592 & 0.508 & \multicolumn{1}{c}{10.7}  \\ \hline 
        \multicolumn{1}{c|}{1} &  0.714 & 0.651 & \multicolumn{1}{c|}{18.8}  & 0.615 & 0.524 & \multicolumn{1}{c}{13.5}  \\ \hline 
        \multicolumn{1}{c|}{2} &   0.728 & 0.684 & \multicolumn{1}{c|}{19.2}   & 0.647 & 0.566 & \multicolumn{1}{c}{13.7}  \\ \hline

      \end{tabular}
      }
      \caption{Results of BLI score on CLWEs using pseudo corpus generated from different quality UMTs.}
      \label{table:bli_per_quality}
    \end{center}

  \end{table}

\subsection{Effect of UMT quality}
\label{subsec: umt_quality}
To investigate the effect of UMT quality on our method, we compare the accuracy of BLI on the CLWEs using pseudo data generated from UMT models of different qualities.
As a translator with low performance, we prepare models that perform fewer iterations on back-translation (BT).
Note that we compare the results on the source-side (English) extension, where the quality of the translation is notably different.
As shown in Table \ref{table:bli_per_quality}, we find that the better the quality of generated data, the better the performance of BLI.

  \begin{table*}[htbp]
    \begin{center}
     \scalebox{0.65}{
      \begin{tabular}{c| c| c | c| cc} \hline
      \multicolumn{2}{c|}{Extension} & \multirow{2}{*}{\begin{tabular}[c]{@{}c@{}}en - fr\end{tabular}} &  \multirow{2}{*}{\begin{tabular}[c]{@{}c@{}}en - de\end{tabular}} &  \multirow{2}{*}{\begin{tabular}[c]{@{}c@{}}en - ja\end{tabular}} \\ \cline{1-2}
        pseudo & parallel &  &   &   \\ \hline 
       - & - & $\textrm{0.621}$ / $\textrm{711}$& $\textrm{0.502}  $ / $\textrm{877}$& $\textrm{0.426}$ / $\textrm{1776} $
        \\ \hline \hline
       $\times$ & $\times$ & $\textrm{0.630} $ / $\textrm{838} $ & $\textrm{0.509} $ 
        / $\textrm{1714} $ & $\textrm{0.429} $ / $\textrm{2301} $
        \\ \hline
     $\checkmark$   & $\times$ & $\textrm{0.686}$ / $\textbf{123} $ & $\textrm{0.569} $  / $\textrm{272} $ & $\textrm{0.454} $ / $\textrm{1050} $
        \\ \hline
      $\checkmark$ &  $\checkmark$ & $\textbf{0.695} $ / $\textrm{144} $& $\textbf{0.585}  $ / $\textbf{183} $ & $\textbf{0.459}$ / $\textbf{1024} $
        \\ \hline

      \end{tabular}
      }
      \caption{Results of BLI score and eigenvector similarity. In each cell, the left cell shows the result of BLI, and the right cell shows the result of eigenvector similarity. Each row indicates, from top to bottom, no extension, extension with non-pseudo data, extension with non-parallel pseudo data,  and extension with parallel pseudo data.}
      \label{table:bli_result2}
    \end{center}
  \end{table*}

 \begin{table*}[htbp]
    \begin{center}
      \scalebox{0.65}{ 
      \begin{tabular}{c| c | c| cc} \hline
        Corpus & fr-A &  de-A &  ja-A \\ \hline \hline
        en & $\textrm{0.621} $ / $\textrm{711}$     & $\textrm{0.502} $ / $\textrm{877}$  & $\textrm{0.426} $ / $\textrm{1776} $
        \\ \hline
        en + pseudo (fr-B)& $\textbf{0.686}  $ / $\textbf{123} $ & $\textrm{0.516}  $ / $\textrm{315} $ & $\textrm{0.421} $ / $\textrm{2194} $
        \\ \hline
        en + pseudo (de-B)& $\textrm{0.621} $ / $\textrm{193} $  & $\textbf{0.569} $ / $\textbf{272} $& $\textrm{0.423}$ / $\textrm{2173} $
        \\ \hline
        en + pseudo (ja-B)&$\textrm{0.568} $ / $\textrm{279} $  & $\textrm{0.454} $ / $\textrm{625} $ & $\textbf{0.454}$ / $\textbf{1050} $
        \\ \hline
      \end{tabular}
      }
           \caption{Results of BLI score and eigenvector similarity. Note that {\tt lang}-A and pseudo ({\tt lang}-B) are not parallel.}
           \label{table:bli_result1}

    \end{center}
  \end{table*}

\subsection{Effect of sharing content}
\label{subsec: content}
In the mapping method, word embeddings are independently trained by monolingual corpora that do not necessarily have the same content. As a result, the difference in the corpus contents can hurt the structural similarity of the two resulting embedding spaces.
We hypothesize that using synthetic parallel data which have common contents for learning word embeddings leads to better structural correspondence, which improves cross-lingual mapping.


To verify the effect of sharing the contents using parallel data, we compare the extensions with a parallel corpus and a non-parallel corpus.
More concretely, we first split the original training data of the source and target languages evenly (each denoted as Split A and Split B). As the baseline, we train CLWEs with Split A. We use the translation of Split A of the target language data for the parallel extension of the source data, and Split B for the non-parallel extension. Also, we compare them with the
extension with non-pseudo data, which is simply increasing the amount of the source language data by raw text.

Along with the BLI score, we show eigenvector similarity, a spectral metric to quantify the structural similarity of word embedding spaces \cite{Sogaard:2018vo}.
To compute eigenvector similarity, we normalize the embeddings and construct the nearest neighbor graphs of the 10,000 most frequent words in each language. We then calculate their Laplacian matrices $L1$ and $L2$ from those graphs and find the smallest $k$ such that the sum of the $k$ largest eigenvalues of each Laplacian matrices is $<90\%$ of all eigenvalues. Finally, we sum up the squared differences between the $k$ largest
eigenvalues from $L1$ and $L2$ and derive the eigen similarity. Note that smaller eigenvector similarity values mean higher degrees of structural similarity. 

Table \ref{table:bli_result2} shows the BLI scores and eigenvector similarity in each extension setting.
The parallel extension method shows a slightly better BLI performance than the non-parallel extension.
This supports our hypothesis that parallel pseudo data make word embeddings space more suitable for bilingual mapping because of sharing content.
In eigenvector similarity, there is no significant improvement between the parallel and non-parallel corpora. This is probably due to large fluctuations in eigenvector similarity values.
Surprisingly, the results show that augmentation using pseudo data is found to be much more effective than the extension of the same amount of original training data.
This result suggests that using pseudo data as training data is useful, especially for learning bilingual models.

\subsection{Effect of reflecting the co-occurrence statistics of the language}
\label{subsec: co-occurrence}

We hypothesize that the translated sentences reflect the co-occurrence statistics of the original language, which makes the co-occurrence information on training data similar, improving the structural similarity of the two monolingual embeddings.

To verify this hypothesis, we experiment with augmenting the source language with sentences translated from a non-target language. To examine only the effect of the co-occurrence statistics of language and avoid the effects of sharing content, we use the extensions with the non-parallel corpus.

Table \ref{table:bli_result1} shows that BLI performance and eigenvector similarity improve with the extension from the same target language, but that is not the case if the pseudo corpus is generated from a non-target language.
These results indicate that our method can leverage learning signals on the other language in the pseudo data.


\begin{table*}[htbp]
    \begin{center}
    \scalebox{0.65}{
      \begin{tabular}{c| ccc | ccc| ccc} \hline
       & \multicolumn{3}{c|}{en-fr} & \multicolumn{3}{|c|}{en-de} & \multicolumn{3}{|c}{en-ja}  \\ \hline
        \multirow{2}{*}{Task} &   \multirow{2}{*}{mapping}  &  \multirow{2}{*}{\begin{tabular}[c]{@{}c@{}}mapping\\(+ pseudo)\end{tabular}}  &  \multirow{2}{*}{\begin{tabular}[c]{@{}c@{}}joint\\training\end{tabular}}  &   \multirow{2}{*}{mapping}  &  \multirow{2}{*}{\begin{tabular}[c]{@{}c@{}}mapping\\(+ pseudo)\end{tabular}}  &  \multirow{2}{*}{\begin{tabular}[c]{@{}c@{}}joint\\training\end{tabular}}   &   \multirow{2}{*}{mapping}  &  \multirow{2}{*}{\begin{tabular}[c]{@{}c@{}}mapping\\(+ pseudo)\end{tabular}}  &  \multirow{2}{*}{\begin{tabular}[c]{@{}c@{}}joint\\training\end{tabular}}   \\
   &  &  &   &  &  &   &  &  &   \\
        \hline \hline
       \multirow{2}{*}{TC}  & $79.5   $ & $  \textbf{82.2}^{\dagger}$ &$ \textrm{79.7}$& $79.0$ &$\textbf{79.3} $ & $\textrm{70.4} $ & $70.4 $ & $\textbf{71.6}^{\dagger} $ &$\textrm{66.7}$ \\
       
       & $(92.6)$ & $(93.3)$ & $(92.5)$& $ (91.7)$& $(92.0)$ & $(91.4)$ & $(92.2)$ & $(93.3)$ &  $(91.9)$
       
       \\ \hline
       \multirow{2}{*}{SA}  & $69.1   $ & $  \textbf{69.5}$ & $\textrm{66.3}$ & $63.7$ &$\textbf{65.1}^{\dagger} $ & $\textrm{62.5} $ & $\textbf{63.5} $ & $62.8 $ &$\textrm{57.3}$ \\
       
       & $(71.8)$ & $(71.9)$ & $(69.9)$& $ (71.1)$& $(70.2)$ & $(70.3)$ & $(70.7)$ & $(70.6)$ &  $(66.8)$
       
       \\ \hline
    
    　 \multirow{2}{*}{DP}  & $63.9   $ & $  \textbf{64.3}$ & $\textrm{64.1}$ & $56.7$ &$\textbf{57.0} $ & $\textrm{55.9} $ & $\textrm{17.8} $ & $\textbf{18.1} $ &$\textrm{17.3}$ \\
       
       & $(73.2)$ & $(73.5)$ & $(75.1)$& $ (73.2)$& $(73.6)$ & $(74.7)$ & $(72.9)$ & $(73.3)$ &  $(74.8)$
        \\ \hline

    　 \multirow{2}{*}{NLI}  & $ 54.4   $ & $  \textbf{54.7}$ & $\textrm{45.0}$ & $55.7$ &$\textbf{56.0} $ & $\textrm{44.7} $ & - & - & -  \\
       
       & $(70.3)$ & $(70.1)$ & $(68.6)$& $ (70.2)$& $(70.3)$ & $(69.7)$ & - & - &  -
       \\ \hline
      \end{tabular}
    }
           \caption{Results of Downstream tasks. Numbers in parentheses indicate the score of English validation data. The scores indicate averages of 20 experiments with different seeds. Statistically significant correlations are marked with a dagger (p $< $0.01).}
           \label{table:d_results}

    \end{center}
  \end{table*}

\section{Downstream Tasks}
\label{sec:DT}


Although CLWEs were evaluated almost exclusively on the BLI task in the past,
\citet{glavas-etal-2019-properly} recently showed that CLWEs that perform well on BLI do not always perform well in other cross-lingual tasks.
Therefore, we evaluate our embeddings on the four downstream tasks: topic classification (TC), sentiment analysis (SA), dependency parsing (DP), and natural language inference (NLI).

\paragraph{Topic Classification}
This task is classifying the topics of news articles. We use the MLDoc \footnote{https://github.com/facebookresearch/
MLDoc} corpus compiled by \citet{SCHWENK18.658}.
It includes four topics: CCAT (Corporate / Industrial), ECAT (Economics), GCAT (Government / Social), MCAT (Markets). As the classifier, we implemented a simple light-weight convolutional neural network (CNN)-based classifier.

\paragraph{Sentiment Analysis}
In this task, a model is used to classify sentences as either having a positive or negative opinion.
We use the Webis-CLS-10 corpus \footnote{https://webis.de/data/webis-cls-10.
html}.
This data consists of review texts for amazon products and their ratings from 1 to 5. We cast the problem as binary classification and define rating values 1-2 as ``negative" and 4-5 as ``positive", and exclude the rating 3. Again, we use the CNN-based classifier for this task.

\paragraph{Dependency Parsing}
We train the deep biaffine parser \cite{Dozat:2017wp} with the UD English EWT dataset\footnote{\url{https://universaldependencies.org/treebanks/en_ewt/index.html}} \cite{Silveira:2014vs}. We use the PUD treebanks\footnote{\url{https://universaldependencies.org/conll17/}} as test data.


\paragraph{Natural Language Inference}
We use the English MultiNLI corpus \cite{williams-etal-2018-broad} for training and the multilingual XNLI corpus for evaluation \cite{conneau-etal-2018-xnli}. XNLI only covers French and German from our experiment.
We train the LSTM-based classifier \cite{bowman-etal-2015-large}, which encodes two sentences, concatenated the representations, and then feed them to a multi-layer perceptron.

In each task, we train the model using English training data with the embedding parameters fixed .
We then evaluate the model on the test data in other target languages.

\subsubsection*{Result and Discussion}

Table \ref{table:d_results} shows the test set accuracy of downstream tasks.  For topic classification, our method obtains the best results in all language pairs. Especially in En-Fr and En-Ja, a significant difference is obtained in Student’s t-test.  For sentiment analysis, we observe a significant improvement in En-De, but cannot observe consistent trends in other languages. For dependency parsing and natural language inference, we observe a similar trend where the performance of our method outperforms other methods, although no significant difference is observed in the t-test. 
The cause of the lower performance of joint-training compared with the mapping method is presumably due to the poor quality of synthetic parallel data as described in § \ref{subsec: bli}.
In summary, given the same amount of data, the CLWEs obtained from our method tend to show higher performance not only in BLI but also in downstream tasks compared with other alternative methods, although there is some variation.



\section{Analysis}

\begin{table*}[htbp]
  \begin{center}
  \scalebox{0.65}{
    \begin{tabular}{c|cc|cc|cccc} \hline
        &  \multicolumn{2}{|c|}{en-fr}   & \multicolumn{2}{|c|}{en-de}  & \multicolumn{2}{|c}{en-ja}  \\ \hline
       corpus & en & fr & en & de & en & ja \\ \hline \hline
      origin &$ 1.60 \times 10^{-3}$ & $1.63 \times 10^{-3} $& $1.51 \times 10^{-3} $&$ 3.78 \times 10^{-3}$& $1.52 \times 10^{-3}$ & $1.03 \times 10^{-3}$
      \\ \hline
      
      pseudo&$ 0.57 \times 10^{-3} $& $0.57 \times 10^{-3}$  & $0.66 \times 10^{-3}$ & $0.59 \times 10^{-3}$  &$ 0.19  \times 10^{-3}$  & $0.17 \times 10^{-3}$
      \\ \hline
    \end{tabular}
    }
    \caption{Type-token ratio of the training corpus (origin) and the pseudo-corpus (pseudo)}
    \label{table:vocab_size}

  \end{center}
\end{table*}

\begin{table}[t]

        \begin{center}
          \scalebox{0.65}{
            \begin{tabular}{cccc} \hline
            corpus & simverb-3500 &  men \\ \hline \hline
            en & $0.259  $& $0.763$
            \\ \hline
            en + pseudo (fr)& $ 0.260 $& $0.767 $
            \\ \hline
            en + pseudo (de)&$ 0.253 $ & $0.768 $
            \\ \hline
            en + pseudo (ja)&$ 0.220 $ & $0.760 $
            \\ \hline
            \end{tabular}
            }
            \caption{Results of word similarity. The scores indicate averages of 3 experiments with different seeds.}
            \label{table:ws_result}

        \end{center}
      \end{table}

\paragraph{Monolingual Word Similarity}

Our method uses a noisy pseudo corpus to learn monolingual word embeddings, and it might hurt the quality of monolingual embeddings.
 To investigate this point, we evaluate monolingual embeddings with the word similarity task. This task evaluates the quality of monolingual word embeddings by measuring the correlation between the cosine similarity in a vector space and manually created word pair similarity.
We use simverb-3500\footnote{http://people.ds.cam.ac.uk/dsg40/simverb.html} \cite{gerz-etal-2016-simverb} consisting of 3500 verb pairs and men\footnote{https://staff.fnwi.uva.nl/e.bruni/MEN} \cite{bruni2014multimodal}
consisting of 3000 frequent words extracted from web text.

Table \ref{table:ws_result} shows the results of word similarity.
The scores of monolingual word embeddings using a French and German pseudo corpus are maintained or improved, while they decrease in Japanese.
This suggests that the quality of monolingual word embeddings could be hurt
due to the low quality of the pseudo corpus or differences in linguistic nature.
Nevertheless, the proposed method improves the performance of En-Ja's CLWE, which suggests that the monolingual word embeddings created with a pseudo corpus have a structure optimized for cross-lingual mapping.


\paragraph{Application to UMT}

UMT is one of the important applications of CLWEs.
Appropriate initialization with CLWEs is crucial to the success of UMT \cite{lample-etal-2018-phrase}.
To investigate how CLWEs obtained from our method affect the performance of UMTs, we compare the BLEU scores of UMTs initialized with CLWEs with and without a pseudo corpus at each iterative step.
As shown in Table \ref{table:blue_results_app}, we observe that initialization with CLWE using the pseudo data result in a higher BLEU score in the first step but does not improve the score at further steps compared to the CLWE without the pseudo data.
\citet{marie-fujita-2019-unsupervised} also demonstrate the same tendency in the CLWE with joint-training.

To investigate this point, we compare the lexical densities of the training corpus and the pseudo-corpus used in the above experiments (§ \ref{sec:ECM}, \ref{sec:DT}) using type-token ratio (Table \ref{table:vocab_size}).
The results demonstrate that the pseudo corpus has a smaller vocabulary per word than the training corpus, and thus it is standardized to some extent as reported in \citet{extentvanmassenhove-etal-2019-lost}.
As a result, specific words might be easily mapped in CLWEs using a pseudo corpus\footnote{In a preliminary experiment, we investigated the variation in performance of cross-lingual mapping with and without pseudo according to the frequency of words in the source language, but there was little correlation between them.}, and then the translation model makes it easier to translate phrases in more specific patterns.
Hence, the model cannot generate diverse data during back-translation, and the accuracy is not improved due to easy learning.

\begin{table}[t]
    \begin{center}
     \scalebox{0.65}{
      \begin{tabular}{c|cc|cc} \hline
      BT & en$ \rightarrow $fr  & fr$ \rightarrow $en & en$ \rightarrow $fr & fr$ \rightarrow $en
       \\
      step &\multicolumn{2}{|c|}{CLWE (no pseudo)} & \multicolumn{2}{|c}{CLWE (+ pseudo)}
       \\ \hline \hline
        0& & \textrm{14.7}&  & \textbf{14.8}
        \\ \hline
        1& \textbf{16.7}& \textbf{18.8}& \textrm{16.1}& \textrm{18.2}
        \\ \hline
        2& \textbf{18.8}& \textbf{19.2}& \textrm{18.2}& \textrm{18.5}
        \\ \hline
        3& \textbf{19.2}& \textbf{19.1}& \textrm{18.6}& \textrm{18.8}
        \\ \hline
      \end{tabular}
      }
      \caption{BLEU scores of UMT at each back-translation step in En-Fr with a phrase
table induced using different CLWEs.}
      \label{table:blue_results_app}
    \end{center}

  \end{table}


\section{Conclusion and Future Work}

In this paper, we show that training cross-lingual word embeddings with pseudo data augmentation improves performance in BLI and downstream tasks.
We analyze the reason for this improvement and found that the pseudo corpus reflects the co-occurrence statistics and content of the other language and that the property makes the structure of the embedding suitable for cross-lingual word mapping.

Recently, \citet{vulic-etal-2019-really} have shown that fully unsupervised CLWE methods fails in many language pairs and argue that researchers should not focus too much on the fully unsupervised settings.
Still, our findings that improve structural similarity of word embeddings in the fully unsupervised setting could be useful in semi-supervised settings, and thus we would like to investigate this direction in the future.

\clearpage
\bibliographystyle{acl_natbib}
\bibliography{acl2021}

\clearpage

\newpage
\section{Appendix}

\appendix

\section{The hyperparameters for downstream tasks}
\label{appendix:DT}

\subsection{Document Classification and Sentiment Analysis}
\begin{table}[h]
\begin{tabular}{clc} \hline
\multicolumn{2}{c}{hyperparameters}                    &            \\ \hline
\multirow{3}{*}{CNN Classifier} & number of filters    & 8       \\
                                & ngram\_filter\_sizes & 2, 3, 4, 5 \\
                                & MLP hidden size      & 32        \\ \hline
\multirow{5}{*}{Training}       & optimizer       & Adam                                                   \\
                                                  & learning rate          & 0.001  \\
                                                  & lr scheduler & halved each time the dev score stops improving\\
                                                  & patience               & 3                                                      \\
                                                  & batch size             & 50             \\ \hline
\end{tabular}
\end{table}

\subsection{Dependency Parsing}
\begin{table}[h]
\begin{tabular}{clc} \hline
\multicolumn{2}{c}{hyperparameters}                           & \multicolumn{1}{r}{}                                   \\ \hline
\multirow{5}{*}{Graph-based Parser}   & LSTM hidden size       & 200                                                    \\
                                     & LSTM number of  layers & 3                                                      \\
                                     & tag representation dim & 100                                                    \\
                                     & arc representation dim & 500                                                    \\
                                     & pos tag embedding dim  & 50                                                     \\ \hline
\multirow{6}{*}{Training}            & optimizer              & Adam                                                   \\
                                     & learning rate          & 0.001  \\
                                     & lr scheduler & halved each time the dev score stops improving\\
                                     & patience               & 3                                                      \\
                                     & batch size             & 32 \\ \hline
\end{tabular}
\end{table}

\subsection{Natural Language Inference}
\begin{table}[h]
\begin{tabular}{clc} \hline
\multicolumn{2}{c}{hyperparameters}                           & \multicolumn{1}{r}{}                                   \\ \hline
\multirow{2}{*}{Sentence Encoder}   & LSTM hidden size       & 300                                                    \\
                                     & LSTM number of  layers & 2                                                    \\ \hline
\multirow{5}{*}{Training}            & optimizer              & Adam                                                   \\
                                     & learning rate          & 0.001 \\
                                     & lr scheduler & halved each time the dev score stops improving\\
                                     & patience               & 3                                                      \\
                                     & batch size             & 64 \\ \hline
\end{tabular}
\end{table}

\end{document}